\documentclass[10pt,twocolumn,letterpaper]{article}

\usepackage{cvpr}              

\usepackage{graphicx}
\usepackage{amsmath} 
\usepackage{amssymb}
\usepackage{booktabs}

\usepackage[pagebackref,breaklinks,colorlinks]{hyperref}
\usepackage{comment}

\usepackage[capitalize]{cleveref}
\crefname{section}{Sec.}{Secs.}
\Crefname{section}{Section}{Sections}
\Crefname{table}{Table}{Tables}
\crefname{table}{Tab.}{Tabs.}

\def\cvprPaperID{1398} 
\def\confName{CVPR}
\def\confYear{2023}

\newcommand{\ourdataset}{UltraStage Dataset }
\newcommand{\ourdata}{UltraStage}
\newcommand{\ourhardware}{PlenOptic Stage Ultra}

\newcommand{\aka}{a.k.a.}

\newcommand{\eqcomma}{\ \ ,}
\newcommand{\eqstop}{\ \ .}

\newcommand{\lightintensity}{L}
\newcommand{\pixelintensity}{g}
\newcommand{\normal}{\mathbf{n}}
\newcommand{\albedo}{\mathbf{a}}

\newcommand{\camerapose}{\mathbf{c}}

\newcommand{\surfacepoint}{\mathbf{x}}
\newcommand{\density}{\delta}

\newcommand{\origin}{\mathbf{o}}
\newcommand{\ray}{\mathbf{r}}

\newcommand{\dir}{\mathbf{d}}

\newcommand{\normalmap}{\mathbf{N}}
\newcommand{\albedomap}{\mathcal{A}}
\newcommand{\materialmap}{\mathcal{M}}
\newcommand{\depthmap}{\mathcal{D}}

\newcommand{\material}{r}

\begin{document}

\title{Relightable Neural Human Assets from Multi-view Gradient Illuminations}



\author{Taotao Zhou$^{1,4}$\thanks{Equal contribution.}
\and
Kai He$^{1*}$
\and
Di Wu$^{1,3*}$
\and
Teng Xu$^{1,4}$
\and
Qixuan Zhang$^{1,3}$
\and
Kuixiang Shao$^{1,4}$
\and
Wenzheng Chen$^{2}$
\and
Lan Xu$^{1}$\thanks{Corresponding author.}
\and
Jingyi Yu$^{1\dag}$
\and \and
$^{1}$ShanghaiTech University \and $^{2}$University of Toronto \and $^{3}$Deemos Technology \and $^{4}$LumiAni Technology
\and
{\tt\small \{zhoutt, hekai, wudi, xuteng, zhangqx1, shaokx, xulan1, yujingyi\}@shanghaitech.edu.cn} \and {\tt\small wenzheng@cs.toronto.edu}
}

\makeatletter
\let\@oldmaketitle\@maketitle
\renewcommand{\@maketitle}{\@oldmaketitle
\vspace{-1.2cm}
\begin{center}
    \centering
    \captionsetup{type=figure}
    \includegraphics[width=\textwidth]{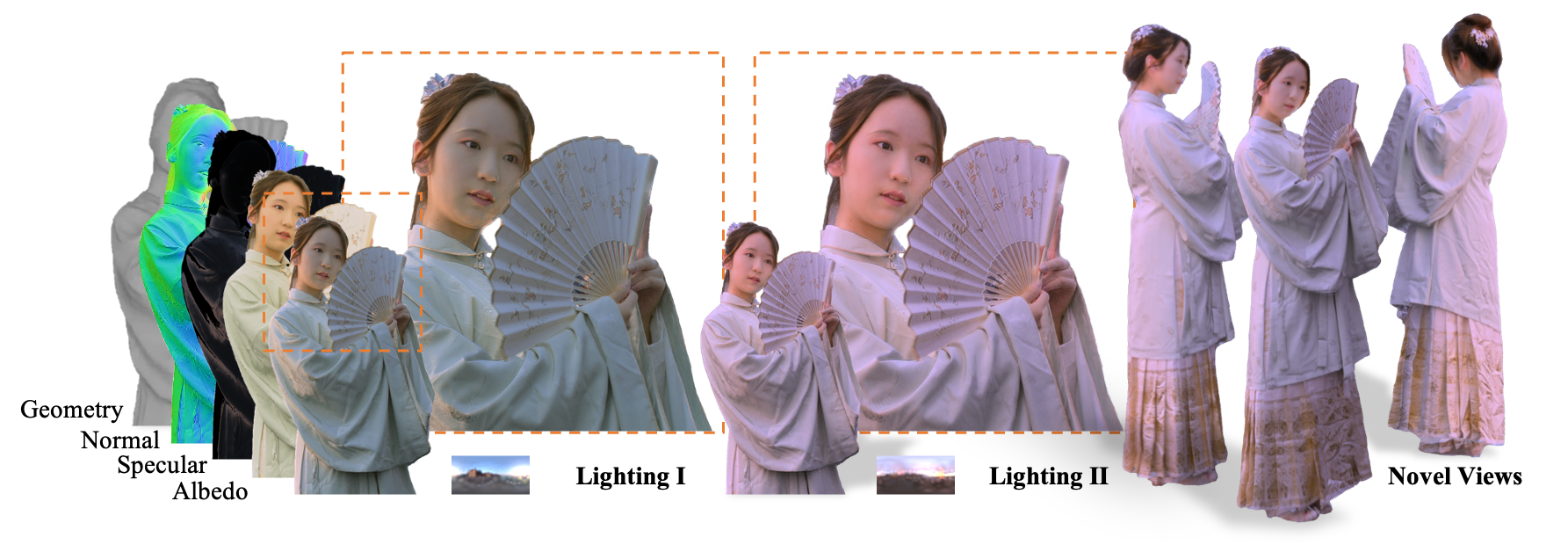}\bigskip
    \vspace*{-1cm}
    \captionof{figure}
    {
    We present {\ourdata}, a new dataset containing more than 2000 human assets captured under multi-view and multi-illumination settings. 
    The high-quality images allow us to extract detailed normal, albedo, and material maps, as well as reconstruct fine geometry (left). 
    We further propose a neural processing pipeline to interpret each capture into a neural human asset, which enables various applications like photo-realistic relighting (middle) and exquisite novel view synthesis (right). 
    Our assets faithfully model human details, e.g., the delicate cloth wrinkles or the vivid classical fan textures.
    }
    \label{fig:teaser}
    \vspace*{-0.2cm}
\end{center}
  }

\makeatother

\maketitle



\begin{abstract}
\vspace*{-0.1cm}
Human modeling and relighting are two fundamental  problems in computer vision and graphics, where high-quality datasets can largely facilitate related research. However, most existing human datasets only provide multi-view human images captured under the same illumination. Although valuable for modeling tasks, they are not readily used in relighting problems. To promote research in both fields, in this paper, we present {\ourdata}, a new 3D human dataset that contains more than 2,000 high-quality human assets captured under both multi-view and multi-illumination settings. Specifically, for each example, we provide 32 surrounding views illuminated with one white light and two gradient illuminations. In addition to regular multi-view images, gradient illuminations help recover detailed surface normal and spatially-varying material maps, enabling various relighting applications. Inspired by recent advances in neural representation, we further interpret each example into a neural human asset which allows novel view synthesis under arbitrary lighting conditions. We show our neural human assets can achieve extremely high capture performance and are capable of representing fine details such as facial wrinkles and cloth folds. We also validate {\ourdata} in single image relighting tasks, training neural networks with virtual relighted data from neural assets and demonstrating realistic rendering improvements over prior arts. {\ourdata} will be publicly available to the community to stimulate significant future developments in various human modeling and rendering tasks. The dataset is available at \href{https://miaoing.github.io/RNHA}{https://miaoing.github.io/RNHA}.
\end{abstract}

\vspace*{-0.3cm}
\section{Introduction}
\label{sec:intro}

Multi-view stereo (MVS) and photometric stereo (PS) have long served as two complementary workhorses for recovering 3D objects, human performances, and environments~\cite{hartley2003multiple, basri2007photometric}. Earlier MVS typically exploits feature matching and bundle adjustment to find ray correspondences across varying viewpoints ~\cite{ingemar1996likelihood, kolmogorov2002multi, triggs1999bundle}  and subsequently infer their corresponding 3D points~\cite{furukawa2009accurate, tola2012efficient, schonberger2016pixelwise}. More recent neural modeling approaches have emerged as a more effective solution by implicitly encoding both geometry and appearance using neural networks~\cite{mildenhall2021nerf}. PS, in contrast, generally assumes a single (fixed) viewpoint and employs appearance variations under illumination changes to infer the surface normal and reflectance (e.g., albedo) ~\cite{nehab2005efficiently, ma2007rapid, basri2007photometric, ma2008facial, fyffe2009cosine, ghosh2009estimating, ghosh2010circularly, ghosh2011multiview, kampouris2018diffuse}. Shape recovery in PS essentially corresponds to solving inverse rendering problems~\cite{ghosh2011multiview} where recent approaches also move towards neural representations to encode the photometric information~\cite{ren2015image, xu2018deep, meka2018lime, meka2019deep}. Most recently, in both MVS and PS neural representations have demonstrated reduced data requirement~\cite{li2020dynamic, Xiang2020OneShotIP} and increased accuracy~\cite{gardner2017learning, georgoulis2017what}. In the context of 3D human scanning, MVS and PS exhibit drastically different benefits and challenges, in synchronization, calibration, reconstruction, etc. For example, for high-quality performance capture, MVS has long relied on synchronized camera arrays but assumes fixed illumination. The most popular and perhaps effective apparatus is the camera dome with tens and even hundreds of cameras~\cite{liu2009point}. Using classic or neural representations, such systems can produce geometry with reasonable quality. However, ultra-fine details such as facial wrinkles or clothing folds are generally missing due to limited camera resolutions, small camera baselines, calibration errors, etc~\cite{wenger2005performance}.

In comparison, a typical PS capture system uses a single camera and hence eliminates the need for cross-camera synchronization and calibrations. Yet the challenges shift to synchronization across the light sources and between the lights and the camera, as well as calibrating the light sources. PS solutions are epitomized by the USC LightStage~\cite{hawkins2001photometric, debevec2002lighting, wenger2005performance} that utilize thousands of light sources to provide controllable illuminations~\cite{ma2008facial, ghosh2011multiview, ma2007rapid, ghosh2009estimating, sun2020light}, with a number of recent extensions~\cite{pandey2021total, guo2019relightables, kampouris2018icl, schwartz2013dome, kim2016single, kampouris2018diffuse, bi2021deep, zhang2021neural, nlt}. A key benefit of PS is that it can produce very high-quality normal maps significantly surpassing MVS reconstruction. Further, the appearance variations can be used to infer surface materials and conduct high-quality appearance editing and relighting applications~\cite{sun2020light, zhang2021neural, bi2021deep, pandey2021total, nlt}. However, results from a single camera cannot fully cover the complete human geometry. Nor have they exploited multi-view reconstructions as useful priors. A unified capture apparatus that combines MVS and PS reconstructions has the potential to achieve unprecedented reconstruction quality, ranging from recovering ultra-fine details such as clothes folds and facial wrinkles to supporting free-view relighting in metaverse applications. In particular, the availability of a comprehensive MVS-PS dataset may enable new learning-based approaches for reconstruction~\cite{mildenhall2021nerf, park2019deepsdf, yao2018mvsnet, xu2021h, su2021nerf, zhao2022humannerf}, rendering~\cite{mildenhall2021nerf, zhang2022modeling, zhang2021physg, verbin2022ref, chen2021dib, wang2021ibrnet}, and generation~\cite{karras2019style, karras2020analyzing, karras2021alias, fu2022stylegan, sarkar2021humangan}. However, there is very little work or dataset available to the public due to challenges on multiple fronts for constructing such imaging systems.

To fill in the gap, we construct {\ourhardware}, an emerging hardware system that conducts simultaneous MVS and PS acquisition of human performances. {\ourhardware} is built upon an 8-meter radius large-scale capture stage with 22,080 light sources to illuminate a performer with controllable illuminations, as well as places 32 cameras on the cage to cover the $360^\circ$ surrounding views of the object. We present detailed solutions to obtain accurate camera-camera and camera-light calibration and synchronization as well as conduct camera ISP correction and lighting intensity rectification. For PS capture, {\ourhardware} adopts tailored illuminations ~\cite{ma2007rapid, guo2019relightables, meka2020deep, meka2019deep}: for each human model, we illuminate it with one white light and two directional gradient illuminations. This produces extra high-quality surface normal and anisotropic reflectance largely missing in existing MVS human reconstruction. A direct result of our system is {\ourdata}, a novel human dataset that provides more than 2,000 human models under different body movements and with sophisticated clothing like frock or cheongsam, to be disseminated to the community.

We further demonstrate several neural modeling and rendering techniques~\cite{mildenhall2021nerf, muller2022instant, zhao2022human, zhang2022modeling, zhang2021physg, chen2021dib} to process the dataset for recovering ultra-fine geometric details, modeling surface reflectance, and supporting relighting applications. Specifically, we show neural human assets achieve significantly improved rendering and reconstruction quality over purely MVS or PS based methods~\cite{campbell2008using, furukawa2009accurate, schonberger2016pixelwise, ghosh2011multiview, fyffe2009cosine, ma2007rapid}, e.g., they can render exquisite details such as facial wrinkles and cloth folds. To use the neural assets for relighting, we adopt albedo and normal estimation networks for in-the-wild human full-body images and we show our dataset greatly enhances relighting quality over prior art~\cite{kanamori2019relighting, tajima2021relighting}. The dataset as well as the processing tools are expected to stimulate significant future developments such as generation tasks in both shape and appearance.
\begin{figure}[t]
  \centering
  \includegraphics[width=1.0\linewidth]{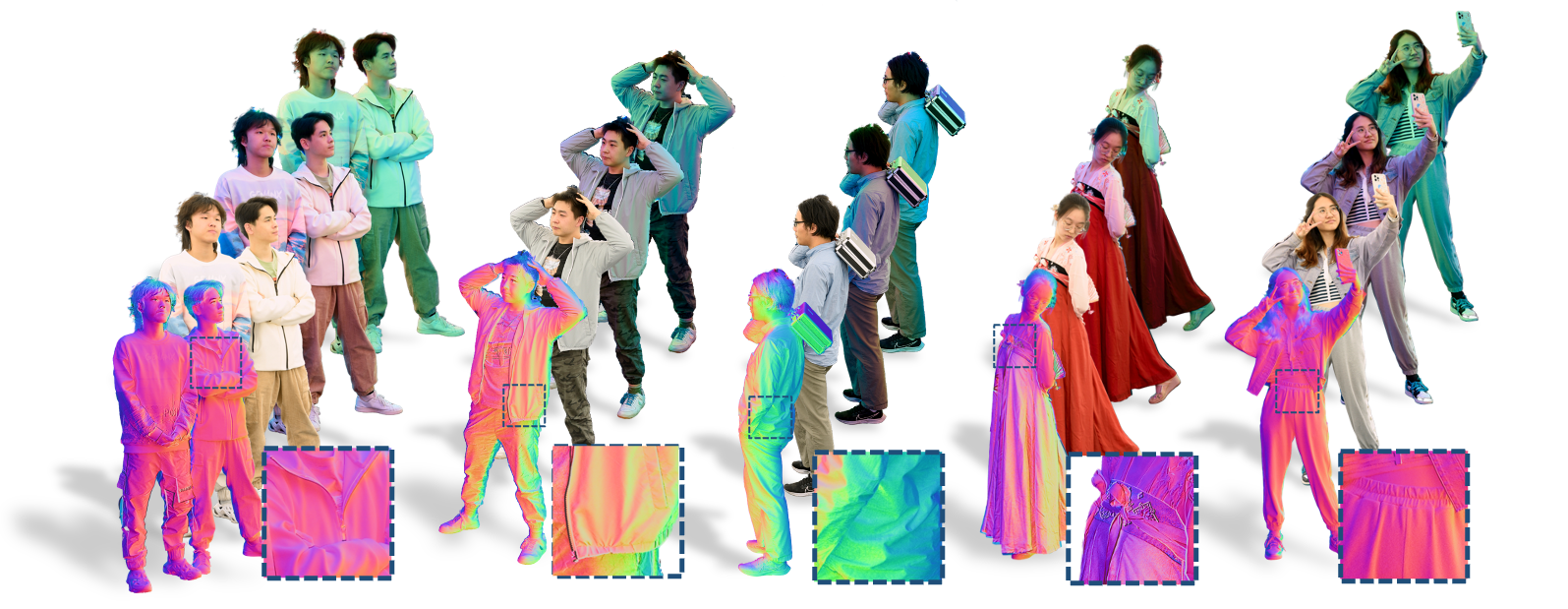}
  \vspace*{-0.8cm}
    \caption{Data overview. 
    	Here we show 5 examples in our dataset. From front to back are the normal map, albedo map, color gradient illumination observation, and inverse color gradient illumination observation, respectively. 
    	In total, {\ourdataset} contains more than 2,000 human-centric scenes of a single person or multiple people with various gestures, clothes and interactions. 
    	For each scene, we provide 8K resolution images captured by 32 surrounding cameras under 3 illuminations. See Supp. for more examples.}
   \label{fig:dataset}
   \vspace*{-0.9cm}
\end{figure}

\begin{figure}[t]
  \centering
  \includegraphics[width=0.95\linewidth]{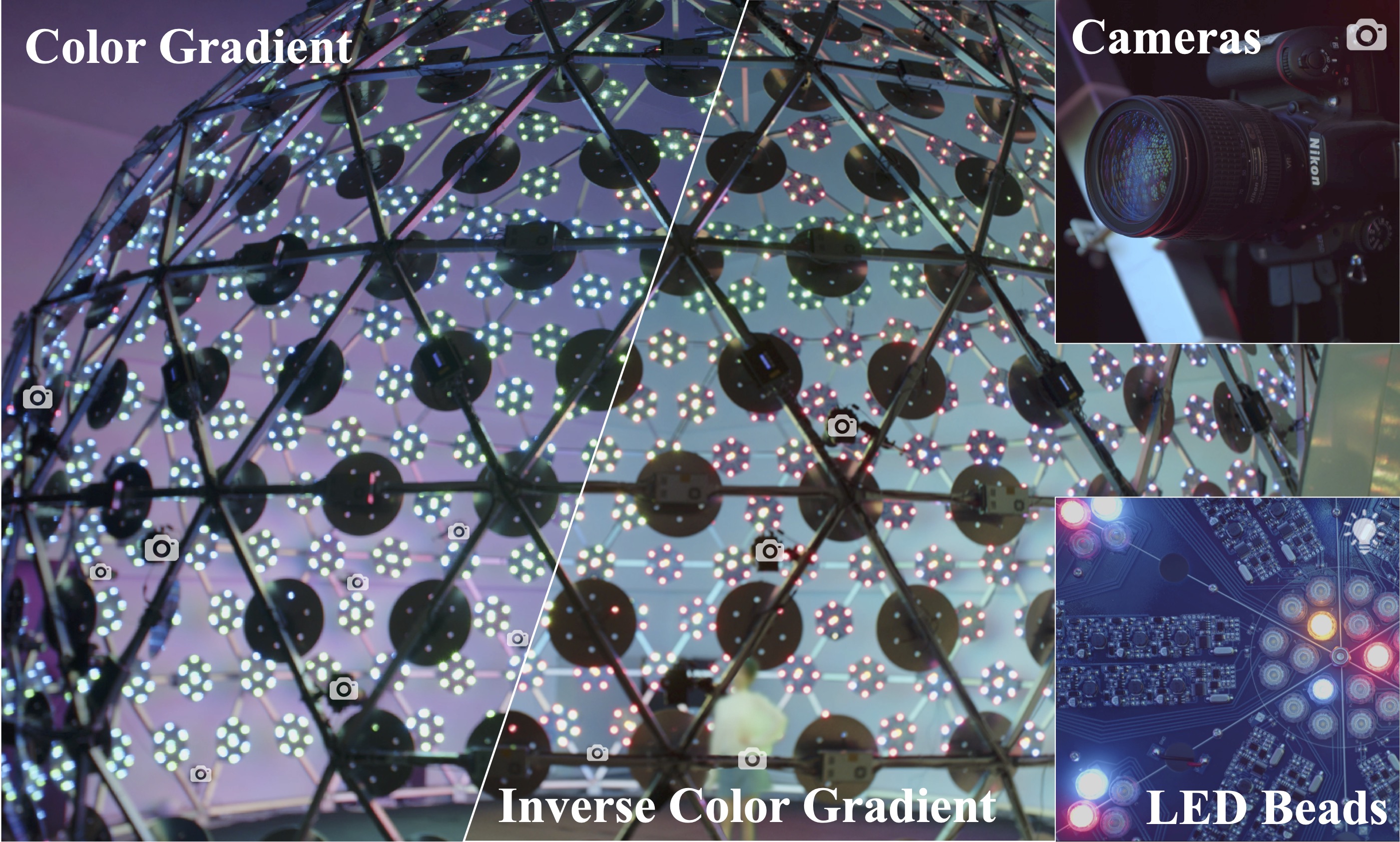}
  \vspace*{-0.3cm}
   \caption{System Overview. The {\ourhardware} is an 8-meter lighting stage composed of 32 sounding cameras and 22,080 controllable light sources.
   It supports both MVS and PS capture settings, enabling high-quality geometry and material acquisition for large-scale subjects and objects. Note that a human is inside in the middle of the picture.
}
   \label{fig:capture_system}
   \vspace*{-0.5cm}
\end{figure}

\vspace*{-0.1cm}
\section{Related Work}
\label{sec:related}

The primary objective of our work is to amalgamate MVS and PS capture techniques into a cohesive system to generate high-quality human performance data. The existing literature on both methods is extensive, but we will focus on discussing those most pertinent to human capture.

\vspace*{-0.3cm}
\paragraph{Multi-view stereo techniques and datasets.}
MVS reconstructs 3D geometry from a set of 2D images captured at different viewpoints \cite{hartley2003multiple,seitz2006comparison,goesele2006revisited} successfully recovering human body~\cite{newcombe2015dynamicfusion, collet2015high, dou2016fusion4d}, face~\cite{klaudiny2012high, lombardi2021mixture,chen2019photo}, clothing~\cite{halimi2022garment}, hair~\cite{nam2019strand}, etc. Earlier works rely on correspondence matching~\cite{ng2003sift, dalal2005histograms, rublee2011orb} and rich textures, making them vulnerable to bare skin and textureless clothing (e.g., dark pants). With the support of deep learning, recent works employ neural representations and differentiable rendering in the MVS pipeline ~\cite{mildenhall2021nerf,muller2022instant,wang2021neus,chen2021mvsnerf} where geometry, appearance, and surface reflectance can be effectively encoded into a tailored neural network~\cite{zhang2022modeling, zhang2021physg, chen2021dib, munkberg2022extracting}.

In human scanning, multi-view capture systems for faces are prevalent due to reduced space, camera, and calibration requirements. However, most efforts focus on static geometry, as synchronizing video cameras is challenging and costly. Several valuable datasets~\cite{marcard2018recovering, li2021ai, Cai2022HuMManM4} have been made available, fostering algorithm development.
MVS reconstruction has inherent limitations, including reduced quality from low resolution, calibration errors, and small camera baselines, struggling to recover fine geometric details. Additionally, MVS typically necessitates fixed and general ambient illumination, leading to less appealing textures.

\vspace*{-0.5cm}
\paragraph{Photometric stereo solutions.}



PS has gained traction as a prominent alternative to MVS for shape reconstruction. PS captures images from a fixed viewpoint and derives per-pixel surface normal maps by analyzing intensity changes under varying illumination. PS techniques rely on normal integration~\cite{horn1986variational} rather than directly producing 3D geometry.

The USC LightStage ~\cite{ghosh2011multiview, ma2007rapid} is a prime example of PS success used in award-winning films. Multiple generations of the LightStage~\cite{ma2007rapid, guo2019relightables, meka2019deep, meka2020deep, zhang2022video} use gradient illumination to obtain normal maps at high efficiency and use coarse geometry as the boundary for normal integration. This geometry may originate from MVS~\cite{ghosh2011multiview, cao2018sparse} or parametric models~\cite{loper2015smpl}. 
{\ourdata} proposes a novel neural modeling pipeline for human asset reconstruction, while The Relightables~\cite{guo2019relightables} uses expensive depth cameras and applies Poisson reconstruction.
One-Light-at-A-Time (OLAT) is an alternative to gradient illumination in PS, requiring ultra-fast camera synchronization with lights. Beyond recovering normal maps, OLAT enables photo-realistic relighting directly~\cite{pandey2021total, sun2020light, zhang2021neural}. Zhang \etal~\cite{nlt} learns a 6D neural light transport function for conducting real-time portrait relighting. Total Relighting~\cite{pandey2021total} produces a photo-realistic relighting effect by the detail normal and albedo, and using the Phong model as a prior.

However, PS systems based on either gradient illumination or OLAT tend to be expensive to construct, especially when combined with MVS. By now the only handful MVS-PS integrated systems are rather small in scale and are not for full-body human captures. 
%
Recent neural approaches~\cite{kaya2022uncertainty, yang2022ps} combine MVS and PS for reconstruction, focusing on static objects under directional lighting. In contrast, our approach leverages gradient illumination for efficient human performance capture and directly utilizes PS normal and albedo to enhance realism.
%
Further, very few datasets have been publicly available whereas we set out to construct such a full-body capture system and produce rich data for the community. 

\vspace*{-0.6cm}
\paragraph{Neural human assets.}
Our research aims to generate neural representations from MVS-PS human capture results and provide raw data. 
While initial Neural Radiance Field (NeRF)~\cite{mildenhall2021nerf} was under the MVS setting, recent advances integrate illuminations ~\cite{srinivasan2021nerv, zhang2021nerfactor, boss2021nerd, boss2021neural}. Srinivasan \etal~\cite{srinivasan2021nerv} trained a neural field from multi-view images under known varying illuminations for free-view rendering and relighting. Most works focus on human faces, with techniques like rendering human eyes with exceptional realism~\cite{li2022eyenerf} and modeling light interactions for high-quality portrait relighting~\cite{sun2021nelf}.


Neural human assets have enabled applications like single-image portrait relighting by inferring geometry, albedo, normal, and other attributes from images. To date, in-the-wild human full-body relighting techniques ~\cite{kanamori2019relighting, tajima2021relighting, lagunas2021single, ji2022geometry} predominantly rely on synthetic training data, using diffuse or simple parametric surface reflectance models, resulting in reduced realism. Even so, publicly available multi-view, multi-lighting human datasets are extremely scarce, and existing ones ~\cite{ji2022geometry,tajima2021relighting,lagunas2021single} contain limited varieties in human subjects, movements, clothing, and other factors.
{\ourdata} contains 2,000 human models engaging in various activities, such as walking, running, standing, stretching, working out, and taking selfies. The dataset also encompasses diverse clothing styles, including classical attire, sportswear, and casual outfits. We captured scenes of individuals interacting with objects, as well as multi-person scenarios.
\section{Data Acquisition}
\label{sec:dataset}

\subsection{System Setup}
\label{sec:dataset:setup}

\begin{table}[t]
\centering
\resizebox{0.47\textwidth}{!}{
\begin{tabular}{cccccc}
\hline \hline
 Hardware & \#Camera & \#Light & Radius & Resolution & \begin{tabular}[c]{@{}c@{}}Color \\light\end{tabular} \\ \hline
\hline
 Wenger \etal~\cite{wenger2005performance} & 1 & 468 & 1m & 1K & - \\ \hline
 Kampouris \etal~\cite{kampouris2018diffuse} & 1 & 336 & 1.25m & - & \checkmark \\ \hline
Guo \etal~\cite{guo2019relightables} & 90 & 21k & - & 4K & \checkmark  \\ \hline
Peng \etal~\cite{peng2021neural} & 21 & - & - & - & - \\ \hline
Bi \etal~\cite{bi2021deep} & \textbf{140} & 460 & 1.1m & 4K & - \\ \hline
\textbf{Ours} & 32 & \textbf{22k} & \textbf{4m} & \textbf{8K} & \checkmark \\ \hline
\end{tabular}
}
\vspace*{-0.3cm}
\caption{Comparison of {\ourhardware} and other hardware for capturing relightable data and human full-body data. Our hardware has advantages over the number of lights (\#lights), radius, and image resolution.}
\label{table:hardware}
\vspace*{-0.7cm}
\end{table}

Our goal is to design an innovative system to acquire high-quality geometry and material properties for large-scale subjects under both MVS and PS capture settings. We construct {\ourhardware}, a giant light stage with an 8-meter diameter, comprising 460 panels with 48 LED beads each, totaling 22,080 individually controllable light sources for versatile illumination. The LEDs support six colors (RGBWAC) for a comprehensive color spectrum. We employ 32 Nikon D750 with Nikkor 24-120mm F4 lenses, arranged $360^\circ$ around the subject, synchronized with the lighting at 5fps. ~\Cref{fig:capture_system} illustrates our system and we compare {\ourhardware} with other similar hardware in ~\cref{table:hardware}.

Precise geometric and photometric calibrations are crucial for shape recovery and post-processing. We devise a method to configure the system, localizing LED bead positions using an alt-azimuth mount, and then calibrate the cameras' intrinsic and extrinsic parameters under the same coordinate system. For photometric calibration, we adjust the camera tone mapping with a color card to output data under linear sRGB color space and rectify the lighting using a method based on LeGendre \etal~\cite{legendre2016practical} to reliably reproduce ambient illumination with RGBWAC light. See supplementary materials for detailed procedures.

\begin{figure}[t]
  \centering
  \includegraphics[width=1.0\linewidth]{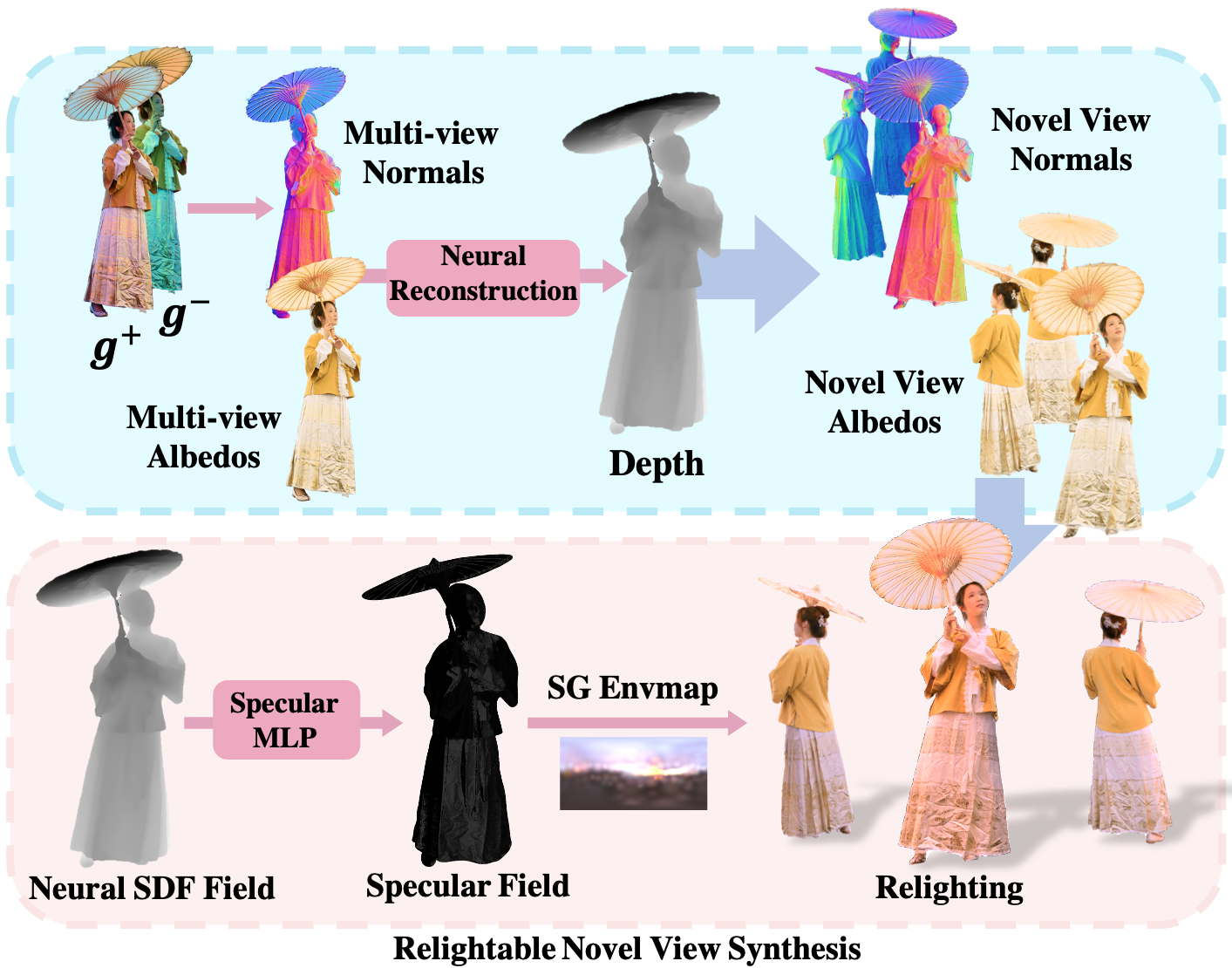}
  \vspace*{-0.8cm}
  \caption{Our pipeline for generating high-quality relightable neural human assets. Given the multi-view input, we first extract corresponding PS normal maps and then train a Neural SDF field from these normal maps (top left). We further adopt depth-guided neural texturing to synthesize novel-view normal and albedo maps (top right) and optimize a neural material field (bottom left) for photo-realistic relighting tasks(bottom, right).}
  \label{fig:pipeline}
  \vspace*{-0.5cm}
\end{figure}

\subsection{Multi-view Normal and Albedo Recovery \\ with Gradient Illuminations}
\label{sec:dataset:capture}

\paragraph{Normal estimation.} 
We capture multi-view images under 3 illuminations, two gradient illuminations and one white light. We record the pixel values captured under two gradient illuminations as $\pixelintensity^+$ and $\pixelintensity^-$. For details on the gradient illuminations, please refer to supplementary materials. Assuming a Lambertian surface BRDF, we follow the approach by Guo \etal~\cite{guo2019relightables} to compute the surface normal $\normal$ as:
\begin{equation}
	\mathbf{d} = \frac{\pixelintensity^+ - \pixelintensity^-}{\pixelintensity^+  + \pixelintensity^-} \eqcomma  \normal= \frac{\mathbf{d}}{|\mathbf{d}|}
	\label{eq:normal} \eqstop
\end{equation}

Here, the normal maps are computed in a ``world'' coordinate, aligned with the camera and lighting system. We compute normal maps for all 32 views, which can then be fused to recover the complete normal map of the human geometry. Several sample normal maps are shown in the Teaser~\cref{fig:teaser} and ~\cref{fig:dataset}.
For a more detailed explanation of gradient illuminations, please refer to ~\cite{ma2007rapid, fyffe2009cosine, guo2019relightables}.

\paragraph{Albedo estimation.} 

Gradient illuminations allow for joint estimation of the surface normal and reflectance parameters, such as albedo or specular. Most recent approaches have been focused on using GI for 3D face captures ~\cite{ma2007rapid,kampouris2018diffuse}. However, the human face has a much simpler, almost convex geometry. In contrast, clothed humans exhibit more complicated geometry with severe self-occlusions such as occluded limbs or clothes folds. Moreover, applying two color gradient illuminations to jointly estimate pixel-wise normal and reflectance parameters is highly ill-posed. We discovered that directly transferring empirical equations from faces to clothed humans results in poor results, where the inferred albedo suffers severe issues from occlusions. See supplementary materials for more analysis.

We define $\lightintensity^0$ as the maximum lighting intensity. $\lightintensity=\lightintensity^0$ means we set the capturing illumination $\lightintensity$ to be the maximum lighting $\lightintensity^0$, \aka, the white light. We also record the pixel value captured under white light as $\pixelintensity^0$ and set the surface albedo $\albedo$ to be $\pixelintensity^0$, details are included in supplementary materials. We do this with two observations: 1) White light helps produce minimum shadows on the human body, largely reducing the influence of self-occlusions. 2) In gradient illuminations normal and albedo are entangled, whereas images captured in white light preserve most albedo information. Although shading is still backed in the albedo, practically, we find it works quite well in all our experiments. Please see \cref{fig:teaser} and \cref{fig:dataset} for examples. For other surface reflectance parameter estimation, we solve it with neural representation (\cref{sec:method}).

\subsection{Dataset Description}
\label{sec:dataset:data}

  \begin{table}[]
  \centering
  \resizebox{0.47\textwidth}{!}{%
    \begin{tabular}{ccccccc}
  \hline \hline
  Dataset & \#Frm & \#Subj & \#View & Res & Relightable & Normal \\ \hline \hline
  3DPW\cite{marcard2018recovering} & 51k & 7 & 1 & - & - & - \\ \hline
  AIST++\cite{li2021ai} & 10M & 30 & 9 & - & - & - \\ \hline
  HuMMan\cite{Cai2022HuMManM4} & \textbf{60M} & \textbf{1k} & 12 & 1K/4K & - & - \\ \hline
    ICT-3DRFE\cite{stratou2011effect}  & $\sim$14k & 23 & 2 & 1K & \checkmark & \checkmark \\ \hline
    Dynamic OLAT\cite{zhang2021neural} & 603k & 36 & 1 & 4K & \checkmark & \checkmark \\ \hline
  \textbf{UltraStage(Ours)} & 192k & 100 & \textbf{32} & \textbf{8K} & \checkmark & \checkmark \\ \hline
  \end{tabular}%
  }
\vspace*{-0.3cm}
\caption{Comparisons of \ourdataset and other published datasets. \ourdataset has a competitive scale in terms of relighting, number of frames (\#Frm), number of viewpoints (\#View), and image resolution (Res); We also contain multi-view normal maps (Normal), and can be used for relighting tasks (Relightable).}
\vspace*{-0.5cm} 
  \label{table:dataset}
  \end{table}

\ourdataset provides a comprehensive multi-view, gradient illumination based full-body human dataset. A unique feature of our capture dome is the adamant space within which a subject can move. Therefore, we manage to acquire data that consists of single and multiple subjects as well as human subjects interacting with objects.
%
%
In total, {\ourdata} provides more than 2,000 human actions, each containing 32 high-resolution 8K images captured under three illuminations, resulting in a total of 192,000 high-quality frames. We ensured a diverse participant pool, recruiting approximately 100 subjects with a balanced gender distribution. {\ourdata} features individuals from a variety of ethnic backgrounds, primarily consisting of Asians, but also representing Caucasians, Africans, Middle Easterners, and other ethnicities. Approximately one-fifth of the participants are middle-aged individuals, while the remaining participants are younger. Each subject performs around 20 poses, with each pose captured under three illumination patterns.


~\Cref{table:dataset} compares \ourdataset and most existing human datasets. While previous works mainly focus on capturing diverse human poses, multi-view human images, and videos prepared for motion capture and human reconstruction tasks. Differently, {\ourdata} captures human-centric images under color gradient illuminations, capable of estimating high-quality normal and reflectance maps that are helpful for both high-quality image-based rendering and geometry reconstruction. We compare our relightable human dataset with other human datasets or relighting datasets in~\cref{table:dataset}. Existing publicly available relightable datasets are all face datasets, such as ICT-3DRFE~\cite{stratou2011effect} with two cameras and single-view Dynamic OLAT~\cite{zhang2021neural}, with one or two cameras. Human datasets like~\cite{marcard2018recovering, Cai2022HuMManM4, li2021ai} contain large-scale frames but do not support relighting. {\ourdata} is the first relightable human-centric dataset with 32 viewpoints and the highest resolution.
\section{Neural Asset Generation}
\label{sec:method}

\begin{figure}[t]
  \centering
  \includegraphics[width=1.0\linewidth]{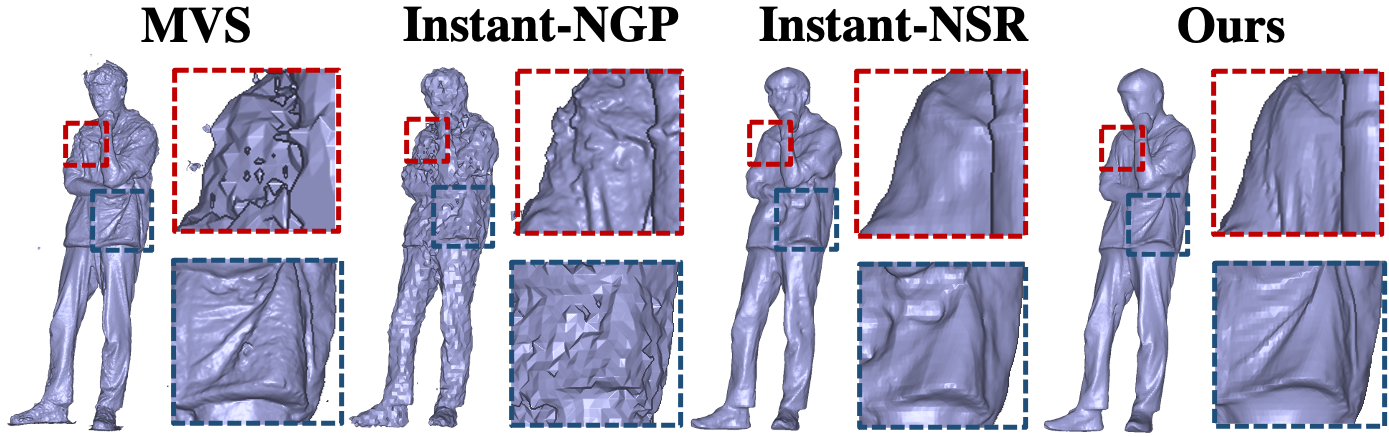}
  \vspace*{-0.8cm}
   \caption{Qualitative comparison on the geometry reconstruction. 
   	We compare our method with traditional MVS solution, Instant-NGP~\cite{muller2022instant}, Instant-NSR~\cite{zhao2022human}.
   	Our method applies PS normal maps to guide geometry generation, demonstrating detailed reconstruction effects.}
   \label{fig:geometry}
   \vspace*{-0.5cm}
\end{figure}

{\ourdata} presents high-quality normal and albedo maps under fixed viewpoints. Inspired by recent advances in neural representations, we further model each example with neural networks, turning it into a neural human asset that enables relightable novel view synthesis.

As shown in Fig.~\ref{fig:pipeline}, our neural processing pipeline consists of two stages. In the first stage, we take the high-quality normal maps as guidance, training a signed distance field (SDF) ~\cite{wang2021neus, zhao2022human} to represent human geometry. We demonstrate in \cref{fig:geometry} how normal priors significantly improve geometry quality, \eg, faithfully reconstructing details such as folds and wrinkles on clothing.

In the second stage, we adopt deferred rendering techniques that can effectively synthesize free-view renderings of the subject under new and more sophisticated lighting environments. We first prepare a set of G-buffers, including normal, albedo, and material maps. To fully leverage our high-quality PS normal and albedo maps, we devise a depth-guided texture blending method, akin to ~\cite{zhao2022human}, to synthesize more detailed albedo and normal buffers and apply inverse rendering frameworks~\cite{zhang2022modeling} to generate material buffer. Once completed, we can shade the image with any desired lighting, which we elaborate on in Sec.~\ref{sec:religting}.

\subsection{High-quality Neural Geometry Modeling}

Given multi-view gradient illumination images, we first extract high-quality normal maps, as described in Sec.~\ref{sec:dataset:capture}. Since normal directions are defined in world coordinates, they are consistent across different views. Following Zhao \etal~\cite{zhao2022human}, we train an SDF field~\cite{wang2021neus} with hash encoding~\cite{muller2022instant} to represent the geometry, where: 
$F_{SDF}: \mathbf{x} \mapsto s $ that maps each 3D location $\mathbf{x} \in \mathbb{R}^3$ to its Signed Distance (SD) value $s \in \mathbb{R}$.
 
However, instead of supervising with RGB images, we feed the normal maps as input to the network. Compared to RGB values which entangle geometry, material, and lighting together, normal maps solely represent surface orientations and thus provide stronger cues for the underlying surface geometry optimization, demonstrating better reconstruction effects.

In ~\cref{fig:geometry}, we compare our neural geometry generation with a traditional MVS solution provided by Agisoft Metashape~\footnote{AgiSoft PhotoScan Professional (Version 1.8.4) (Software). (2022*). Retrieved from http://www.agisoft.com/downloads/installer/} and two recent learning-based neural radiance fields~\cite{muller2022instant, zhao2022human} which take RGB images as input. While we adopt the same network architecture as Zhao \etal~\cite{zhao2022human}, we show normal maps significantly improve the reconstruction quality, preserving more fine-grained details like cloth wrinkles on trousers and dresses.

\begin{figure}
	\centering
	\includegraphics[width=1.0\linewidth]{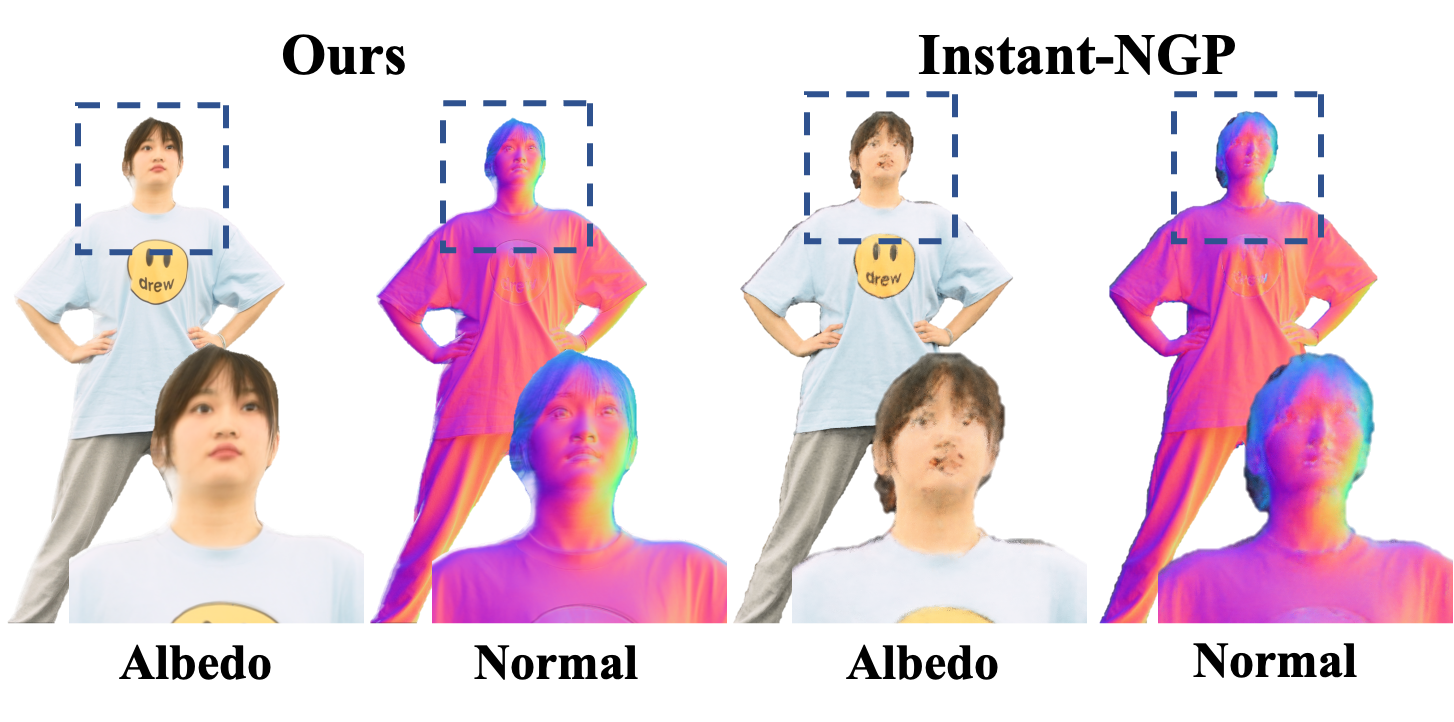}
	\vspace*{-0.4cm}
	\caption{Qualitative comparison on depth-guided G-buffer generation. We compare our method with volume rendering techniques. 
		Specifically, we train normal and albedo neural fields with Instant-NGP~\cite{muller2022instant} and apply volume rendering to generate corresponding G-buffers. By explicitly employing the high-quality PS priors, depth-guided G-buffer generation produces more photorealistic albedo and accurate normal maps.}
	\label{fig:com_NTB}
	\vspace*{-0.6cm}
\end{figure}

\subsection{Relightable Novel View Synthesis}
\label{sec:religting}

By employing deferred rendering techniques, we are able to render neural human assets in arbitrary views and illuminations. 
Specifically, given a new camera pose $\camerapose \in SE(3)$, we first generate corresponding G-buffers, including the normal map $\normalmap \in \mathbb{R}^{h\times w \times 3}$, albedo map $\albedomap \in \mathbb{R}^{h\times w \times 3}$, material map $\materialmap \in \mathbb{R}^{h\times w \times 1}$. To take advantage of the photometric priors, we utilize the pretrained SDF field to synthesize a depth buffer $\depthmap \in \mathbb{R}^{h\times w \times 1}$, then apply reprojection to query corresponding normal and albedo values from the PS normal and albedo maps, details of which we provide later.

With all the prepared G-buffers, we then adopt the general rendering equation (RE)~\cite{kajiya1986rendering} to shade them with desired illuminations. Specifically, for each pixel in the G-buffer, we query its 3D surface location $\mathbf{x}$ from $\depthmap$,  normal $\mathbf{n}$ from $\normalmap$, albedo $\mathbf{a}$ from $\albedomap$, and material $\mathbf{m}$ from $\materialmap$. The outgoing radiance $L_o$ in viewing direction $w_o$ is then computed as:
\begin{equation}
	L_o(w_o; \mathbf{x}) = \int_{\Omega}L_i(w_i; \mathbf{x} )f_r(w_o,w_i; \mathbf{x})\left | w_i\cdot n\right |dw_i \eqcomma
	\label{eq:render_equation}
\end{equation}

where $L_i(w_i; x)$ is the incident radiance at position $\mathbf{x}$ from direction $w_i$ ,  $f_r(w_o, w_i; x)$ is the BRDF that consumes material parameters $\mathbf{a}$ and $\mathbf{m}$, and $\left | w_i\cdot n\right |$ is the cosine foreshortening term.

Although physically correct, the rendering equation contains an integral that has no analytic solution. Traditional graphics pipelines solve it with Monte Carlo methods which are time-consuming and memory-inefficient. 
To accelerate rendering speed, following~\cite{zhang2021physg,chen2021dib} we approximate RE with Spherical Gaussian, \eg, representing the lighting, BRDF, and cosine term with one or more spherical Gaussian components. In the following paragraphs, we elaborate on how to acquire G-buffers and conduct SG approximation to render novel views.

\begin{table}[]
\centering
\resizebox{0.48\textwidth}{!}{%
\begin{tabular}{c|ccc|ccc}
\hline
\multicolumn{1}{l}{} & \multicolumn{3}{c}{Albedo Map} & \multicolumn{3}{c}{Normal Map} \\ \hline
Method & PSNR$\uparrow$ & SSIM$\uparrow$ & RMSE$\downarrow$ & Mean & \textless{}$5^{\circ}$ & \textless{}$25^{\circ}$ \\
Ours   & \textbf{30.868} & \textbf{0.972} & \textbf{0.029} & \textbf{4.296$^{\circ}$} & \textbf{80.848\%} & \textbf{96.293\%} \\
Instant-NGP & 30.529 & 0.968 & 0.030 & 4.314$^{\circ}$ & 79.891\% & 96.013\% \\
\hline
\end{tabular}%
}
\vspace*{-0.3cm}
\caption{Quantitative evaluation on the quality of novel view albedo and normal maps. We separate two views from captured views for testing, and evaluate albedo loss and normal vector error in angular. Our reprojected albedo and normal maps produce higher precision in comparison to volume rendering of Instant-NGP~\cite{muller2022instant}.
}
\label{table:com_albedo_normal}
\vspace*{-0.7cm}
\end{table}

\vspace*{-0.3cm}
\paragraph{Depth-guided G-buffer reprojection.} %

The pretrained SDF field allows us to query the SDF value of any 3D point $\surfacepoint$.  We then following NeuS~\cite{wang2021neus} to convert it into density $\delta$: 

\begin{equation}
	\density(\surfacepoint) = \max(\frac{-\frac{\text{d}\Phi_{s}}{\text{d}t}(f_\text{sdf}(\surfacepoint))}{\Phi_{s}(f_\text{sdf}(\surfacepoint))},0) \eqcomma
\end{equation}
where $\Phi_{s}(x) = \text{Sigmoid}(sx)$, $s$ is a learnable parameter. The volume density field further allows us to render the depth map $\depthmap$ with the standard NeRF volumetric rendering equation~\cite{mildenhall2021nerf}. For example, let $\ray = \origin  + t\dir$ denote the camera ray with origin $\origin$ and direction $\dir$. The alpha-composited depth map $\depthmap$ along the ray can then be estimated as 
\begin{equation}
	\depthmap(\ray)  = \int_{t_n}^{t_f} T(t) \density(\ray(t))t dt,
	\label{eq:vol_rendering}
\end{equation}
where $T(t) = \exp{(-\int_{t_n}^{t}\density(\ray(s))ds)}$ denotes the accumulated transmittance, and $t_n$, $t_f$ are the near and far bound, respectively.

The depth map helps determine the 3D surface location $\surfacepoint$ at each pixel. We then adopt a neural depth-guided reprojection technique~\cite{zhao2022human, eisemann2008floating} to decide its normal $\normal$ and albedo $\albedo$. Specifically, we reproject $\surfacepoint$ into nearby $K$ PS views to query the corresponding high-quality normal and albedo values $\{ \normal_i, \albedo_i \}_{i=1}^{K}$. The normal and albedo in the novel view are computed by the weighted blending of all the queried candidates, where $(\normal, \albedo) = (\sum_{k=1}^{K} w_k \normal_k, \sum_{k=1}^{K} w_k \normal_k)$.

{\ourhardware} has densely surrounded cameras, for each novel view, we set $K=6$ and utilize the six nearest PS views to get the final blending result. Following~\cite{zhao2022human}, we train a blending weight network on Twindom dataset~\footnote{https://web.twindom.com/}, which predicts $K$ weights given all the queried values. The training details can be found in the supplementary materials. Note that although it is possible to train neural networks to predict normal and albedo at any 3D point $\surfacepoint$ and apply volume rendering to synthesize the corresponding normal and albedo maps, in practice, the network tends to produce over-smoothed results that lose sharp details of facial expressions or cloth patterns. In contrast, our depth-guided reprojection directly benefits from the ultra-high quality albedo and normal maps in PS views, resulting in more delicate rendering effects. We show qualitative and quantitative comparison results in ~\cref{fig:com_NTB} and ~\cref{table:com_albedo_normal}, respectively.

\begin{figure}
	\centering
	\includegraphics[width=1.0\linewidth]{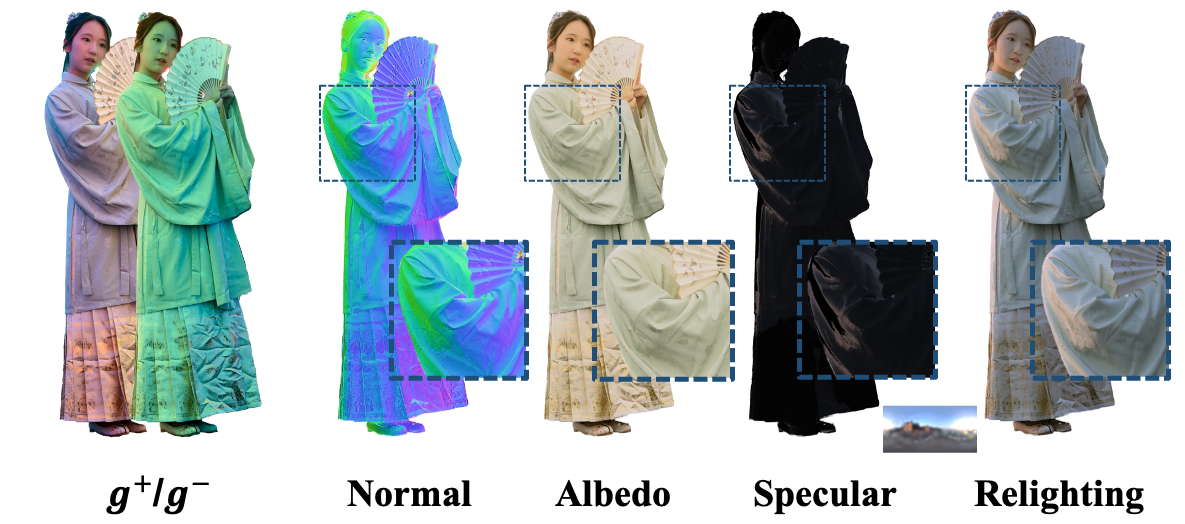}
	\vspace*{-0.8cm}
	\caption{Given color gradient illumination observations $g^+$, $g^-$, by taking high-quality normal and albedo maps as priors, we learn a neural material field on the pre-trained neural geometry surface, enabling photo-realistic relighting results. Note that the specular map is rendered in the target illumination.}
	\label{fig:material}
	\vspace*{-0.5cm}
\end{figure}

\vspace*{-0.3cm}
\paragraph{Material optimization with normal and albedo prior.}

The synthesized normal and albedo G-buffers have admitted nice relighting results under diffuse BRDF settings. We further estimate spatially-varying surface roughness to model view-dependent specular effects. As mentioned earlier, directly applying empirical PS equations to human images fails to get reflectance parameters, potentially due to the complex geometry. Consequently, we cannot use reprojection to acquire the material G-buffer. Instead, we propose optimizing surface material parameters within inverse rendering frameworks, leveraging the prepared surface normal and albedo priors.

Specifically, we follow the state-of-the-art inverse rendering work~\cite{zhang2022modeling} to suit our MVS and PS settings. We optimize material parameters with known illumination (gradient illuminations), geometry (SDF field), normal and albedo (G-buffers). Therefore, we fix all of them and only train a network to predict material, where:
$F_{MAT.}: \mathbf{x} \mapsto \material $
that maps each 3D location $\mathbf{x} \in \mathbb{R}^3$
to its roughness value $\material \in \mathbb{R}$. Following Zhang \etal~\cite{zhang2022modeling}, we set the specular reflectance in the Fresnel term as 0.02. Zhang \etal~\cite{zhang2022modeling} also models the visibility of each point. Here, we assume the shading information has been encoded in albedo maps so we set this term to 1.

During optimization, in the forward process, we first apply volume rendering to produce material G-buffers, similar to the depth buffer generation (\cref{eq:vol_rendering}). We then apply them in general rendering equation (\cref{eq:render_equation}) with Spherical Gaussian approximation~\cite{zhang2022modeling,zhang2021physg,chen2021dib} to synthesize novel view images. In the backward process, we compute the loss between the rendered images and the gradient illumination GT. The optimization details can be found in the supplementary materials. We show the optimized material in \cref{fig:material}.

\vspace*{-0.5cm}
\paragraph{Novel view synthesis and relighting.}

\begin{figure}[t]
  \centering
  \includegraphics[width=1.0\linewidth]{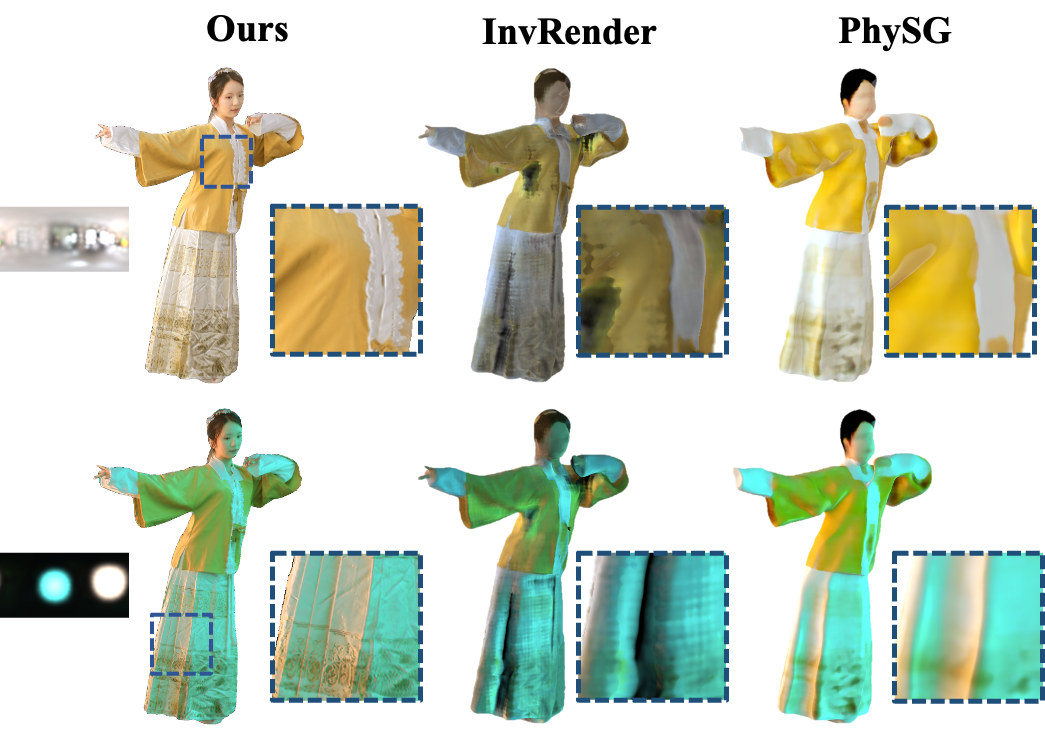}
  \vspace*{-0.8cm}
   \caption{Qualitative comparison on relighting under novel viewpoint with two recent neural relightable novel view synthesis approaches~\cite{zhang2022modeling, zhang2021physg}. Our method takes advantage of photometric priors enabling realistic rendering under complex novel illuminations, with high-frequency details preserved.}
   \vspace*{-0.6cm}
   \label{fig:novel_view_relight}
\end{figure}

With the SDF and material networks, along with high-quality PS normal and albedo maps, now we are able to render each human example in arbitrary viewpoints and illuminations. Thanks to the ultra-fine details in the PS priors, our neural human assets achieve extremely high capture performance, capable of representing fine details such as facial wrinkles and cloth folds. In \cref{fig:novel_view_relight} we compare our novel view synthesis and relighting effects with several baselines, demonstrating significant improvements in rendering quality. We provide more examples in the supplementary materials.
\section{Single Image Relighting}

\begin{figure}[t]
    \centering
    \includegraphics[width=0.7\linewidth]{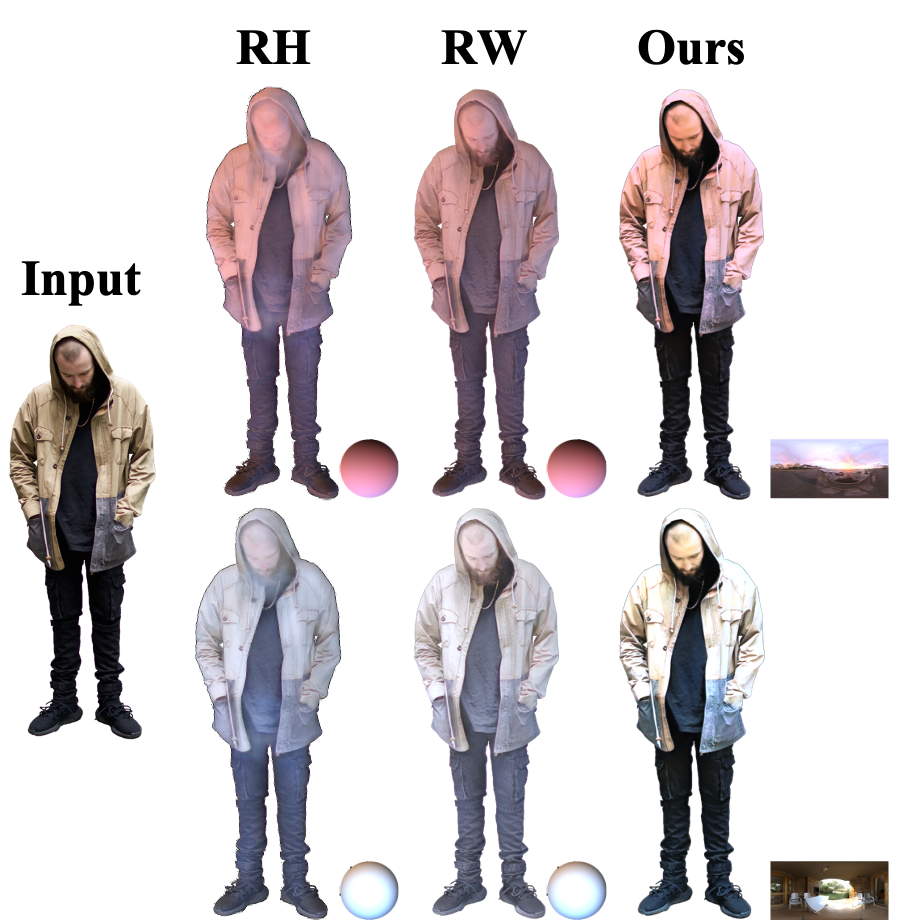}
    \vspace*{-0.3cm}
      \caption{Our relighting results produce sharper details, especially on clothes textures. The reflectance under novel illuminations is more realistic.}
      \label{fig:single_image_relight}
      \vspace*{-0.7cm}
\end{figure}

Our neural human assets allow for novel view synthesis under arbitrary lighting conditions, which can be used to boost a variety of important downstream tasks. In this section, we verify it on the single image relighting task. As a challenging problem, it heavily relies on accurate albedo and normal recovery from the input image. Next, it synthesizes images with new illuminations either from networks~\cite{kanamori2019relighting, tajima2021relighting, lagunas2021single, ji2022geometry} or traditional graphics pipelines. 

Despite recent advancements in the field~\cite{chabert2006relighting, guo2019relightables, kanamori2019relighting,tajima2021relighting, lagunas2021single, ji2022geometry}, photo-realistic human relighting remains challenging, especially for images in the wild.
The primary obstacle is the lack of high-quality training data, \ie, photo-realistic images with GT normal, albedo, and lighting conditions. Consequently, existing human relighting works typically train with synthetic datasets~\cite{kanamori2019relighting, ji2022geometry, lagunas2021single}, leading to significant performance degradation when applied to real-world images. In contrast, {\ourdata} supports synthesizing infinite realistic examples with GT normal, albedo, and lighting labels. We show that even a relatively simple human relighting network, when trained with {\ourdata} dataset, outperforms prior works~\cite{kanamori2019relighting, tajima2021relighting}.

\vspace*{-0.5cm}
\paragraph{Dataset details.}
We first utilize {\ourdata} to synthesize a large-scale dataset for a single image relighting task. Specifically, we select 500 human models with diverse poses and clothing. We also prepare an environment dataset with 2965 HDR illuminations, encompassing both indoor and outdoor lighting conditions from Laval Indoor HDR dataset~\cite{gardner2017learning}, Laval Outdoor HDR dataset~\cite{lalonde2014lighting} and HDRI Haven~\footnote{https://hdrihaven.com/}.
For each person, we render three views under randomly chosen illuminations, creating a dataset with 1500 examples in total. In each example, we prepare a human image together with its corresponding normal and albedo maps.  

\vspace*{-0.5cm}
\paragraph{Network details.}
As our goal is to demonstrate the dataset's capacity, therefore, we choose a very simple network design. Specifically, we employ two U-Net-structure~\cite{ronneberger2015u} networks, named as AlbedoNet and NormalNet, to directly predict albedo and normal maps from the input human image. The networks are supervised with MSE Loss.
We train each network on one A6000 GPU for 24 hours with a learning rate of $10^{-5}$. After training, we then apply networks to arbitrary human images and predict the corresponding normal and albedo maps. Although albedo and normal only admit diffuse relighting effects, we find that the results are visually appealing, thanks to the powerful training dataset. We believe our dataset can facilitate a series of relighting research. 
We will also make this relighting dataset public.

\begin{table}[t!]
	\centering
	\vspace{-3mm} 
 
	\resizebox{0.6\linewidth}{!}{
		\addtolength{\tabcolsep}{6pt}
		\begin{tabular}{l|c}
			\toprule
			& \%  Ours is preferred \\
			\midrule
			vs RH~\cite{kanamori2019relighting} & $96.6$   \%  \\ 
			vs RW~\cite{tajima2021relighting}      	& $83.3$   \%  \\
			\bottomrule
		\end{tabular}
	} 
	\vspace{-3mm} 
	\caption{User study results of human relighting quality. Most users prefer our results over baseline methods.} 
	\label{table:userstudy} 
	\vspace{-8mm} 
\end{table} 

\vspace*{-0.5cm}
\paragraph{Experiments.}

We compare our model with two open-source human relighting methods RW~\cite{tajima2021relighting} and RH~\cite{kanamori2019relighting}, with the former being a following-up work of the latter. We qualitatively compare our relighting results with them in \cref{fig:single_image_relight}, where the image is selected from the testing dataset in~\cite{kanamori2019relighting}. Our model produces more detailed albedo and accurate normal maps, resulting in superior relighting effects, as evidenced by the distinct cloth wrinkles and natural shading. It is worth noting that we achieve these outcomes solely through basic normal and albedo prediction networks. Applying more advanced designs~\cite{peng2021neural} will definitely improve the effects, which we leave for future works.

We further conduct a human study in ~\cref{table:userstudy} on all the testing images in~\cite{kanamori2019relighting}. It contains 6 testing images in total. We relight each one with 5 new illuminations, resulting in 30 examples. Participants are asked to select the more natural-looking relighting, and our approach receives more than 80\% preferences, further attesting to the efficacy of our dataset. More details of the human study can be found in the supplementary materials.
\vspace*{-0.1cm}
\section{Conclusions}
\label{sec:conclusions}

\vspace*{-0.1cm}
We have presented {\ourdata}, a novel 3D human dataset to bridge human modeling and relighting, which contains more than 2,000 high-quality human assets captured under multi-view and multi-illumination settings unseen before. We further interpret this rich captured input into neural human assets, allowing photo-realistic novel-view synthesis under arbitrary lighting conditions. Extensive experiments demonstrate the superiority of {\ourdata}. With the above unique characteristics and rareness of our {\ourdata}, we believe it is critical for future research about high-quality human modeling, with endless potential applications.

\vspace*{-0.5cm}
\paragraph{Acknowledgements.} We thank Hongyang Lin, Qiwei Qiu and Qingcheng Zhao for building the hardware. This work was supported by National Key R$\&$D Program of China (2022YFF0902301), NSFC programs (61976138, 61977047), STCSM (2015F0203-000-06), and SHMEC (2019-01-07-00-01-E00003). We also acknowledge support from Shanghai Frontiers Science Center of Human-centered Artificial Intelligence (ShangHAI).

{\small
\bibliographystyle{ieee_fullname}
\bibliography{egbib}
}

\end{document}